\documentclass[preprint]{elsarticle}

\journal{Neurocomputing}

\usepackage{times}
\usepackage{epsfig}
\usepackage{graphicx}
\usepackage{amsmath}
\usepackage{amssymb}

\usepackage{bm}
\usepackage{bbm}
\usepackage{subfig}
\usepackage{multirow}
\usepackage{enumitem}
\usepackage{comment}
\usepackage{booktabs}

\begin{document}

\begin{frontmatter}

\title{HEMP: High-order Entropy Minimization\\for neural network comPression}

\author[torino,mycorrespondingauthor]{Enzo Tartaglione}
\ead{enzo.tartaglione@unito.it}
\author[parigi]{St\'ephane Lathuili\`ere}
\author[torino,parigi]{Attilio Fiandrotti}
\author[parigi]{Marco Cagnazzo}
\author[torino]{Marco Grangetto}

\address[torino]{University of Torino, Torino, Italy}
\address[parigi]{T\'el\'ecom Paris, Paris, France}

\cortext[mycorrespondingauthor]{Corresponding author}

\begin{abstract}
We formulate the entropy of a quantized artificial neural network as a differentiable function that can be plugged as a regularization term into the cost function minimized by gradient descent. Our formulation scales efficiently beyond the first order and is agnostic of the quantization scheme.
The network can then be trained to minimize the entropy of the quantized parameters, so that they can be optimally compressed via entropy coding. We experiment with our entropy formulation at quantizing and compressing well-known network architectures over multiple datasets.
Our approach compares favorably over similar methods, enjoying the benefits of higher order entropy estimate, showing \emph{flexibility} towards non-uniform quantization (we use Lloyd-max quantization), \emph{scalability} towards any entropy order to be minimized and \emph{efficiency} in terms of compression. We show that HEMP is able to work in synergy with other approaches aiming at pruning or quantizing the model itself, delivering significant benefits in terms of storage size compressibility without harming the model's performance.
\end{abstract}


\end{frontmatter}


\section{Introduction}
\label{sec:introduction}

Artificial Neural Networks (ANNs) achieve state-of-the-art performance in several tasks via complex architectures with millions of parameters.
Deploying such architectures over resource-constrained devices such as mobiles or autonomous vehicles entails tackling a number of practical issues.
Such issues include tight bandwidth and storage caps for delivering and memorizing the trained networks and limited memory for its deployment.
\begin{figure}

    \centering
    \includegraphics{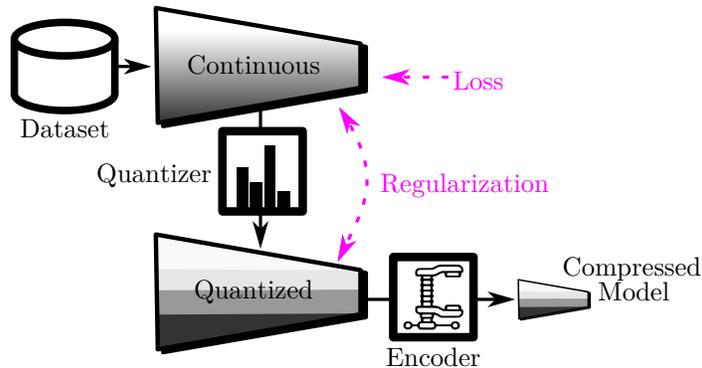}
    \caption{Proposed approach for neural network compression. At training time, we employ two parametrizations of the same neural network: continuous parameters are used for loss minimization, while quantized parameters are used for high-order entropy estimation. A regularization term enforces consistency between the neuron activations of the networks and a low entropy of the quantized network. The final model is obtained using any entropy-based encoder.\label{fig:teaser}}

\end{figure}
\noindent
Let us assume a neural network has to be deployed to a device such as a smartphone or an autonomous car over a wireless link.
Downloading the network may exhaust the subscriber's traffic plan, plus the downloaded network will take storage on the device that will be unavailable to other applications. In an autonomous driving context, safety-critical updates may be delayed due to the limited bandwidth available over the wireless channel~\cite{samarakoon2019distributed}.
Such examples show the importance of efficiently compressing neural networks for transmission and storage purposes.

Multiple approaches have been proposed to compress neural networks.
A first approach consists is designing the network topology from the ground up to encompass fewer parameters~\cite{iandola2016squeezenet, sandler2018mobilenetv2}.
Needless to say, this approach requires designing novel topologies from scratch.
A second approach consists in pruning some parameters from the network, i.e. removing some connections between neurons~\cite{molchanov2017variational, tartaglione2018learning, louizos2017learning}, yielding a sparse topology.
Pruning might reduce the memory footprint~\cite{courbariaux2015binaryconnect, zhou2016dorefa, mishra2017wrpn}, however it does not necessarily minimize storage or bandwidth requirements.
A third approach consists in quantizing the network parameters~\cite{kim2015compression, xu2018deep, wiedemann2019deepcabac}, possibly followed by entropy-coding the quantized parameters.
Similar approaches achieve promising results, however most quantization schemes just aim at learning a compressible representation of the parameters~\cite{wiedemann2019deepcabac, oktay2019scalable, zhou2016dorefa} rather than properly minimizing the compressed parameters entropy.
Indeed, the entropy of the quantized parameters is not differentiable and cannot be easily minimized in standard gradient descent-based frameworks.
In this work, we tackle the problem of compressing a neural network by minimizing the entropy of the compressed parameters at learning time. Enhancing model's compressibility we are able to reduce required bandwidth for bit streaming as well as storage required. Deep models are redundant~\cite{cheng2015exploration, lee2021redundancy}: hence, there is a overhead in the deep model's representation, which can be hereby compressed.
\\
\indent This work introduces HEMP, a method that relies on high-order entropy minimization to allow for efficiently compression of the parameters of a neural network. 
The proposed method is illustrated in Fig.~\ref{fig:teaser}. The main contribution of HEMP is the differentiable formulation of the quantized parameters' entropy, which can be extended beyond first-order with finite computational and memory complexity. Namely, HEMP relies on a twin parametrization of the neural network: continuous parameters and the corresponding quantized parameters, where the entropy of the latters is estimated from the entropy of the former. We design a regularization term around our entropy formulation that can be plugged into gradient descent frameworks to train a network to minimize the entropy of the quantized parameters No assumptions are made on the quantization scheme (including non-uniform quantization), nor entropy coding scheme, that are not part of the proposed method and towards which our method is totally agnostic.\\
Other techniques like $\ell_1$ norm or rank-based ones aim at removing parameters: this has a different effect on the distribution of the parameters. Indeed, while these other approaches maximize the frequency of the pruned parameters, which are still encoded with zeros, HEMP is more general as it is able to enhance the compressibility of any quantized representation for the values.
\\\indent We experiment with different quantization and entropy coding scheme showing that training a network to minimize the 2-nd order entropy of the quantized parameters is already sufficient to outperform state-of-the-art competing schemes.

The rest of the paper is organized as follows. Sec.~\ref{sec:sota} reviews state-of-the-art approaches in network compression, Sec.~\ref{sec:method} introduces the proposed high-order entropy regularizer, and the overall training procedure is described in Sec.~\ref{sec:steps}. Experimental results are discussed in Sec.~\ref{sec:experiments} and finally, in Sec.~\ref{sec:conclusion} conclusions are drawn.

\section{Related works}
\label{sec:sota}

A lot of work has been done around neural network size reduction. In general, we can group them into three large categories, according to their primary goal.\\
\paragraph{Minimizing the architecture} Focusing from the architectural point of view, it is possible to design some memory-efficient deep networks, typically relying on strategies like channel shuffling, point-wise convolutional filters, weight sharing or a combination of them. Some examples of customized deep networks towards memory footprint reduction are SqueezeNet~\cite{iandola2016squeezenet}, ShuffleNet~\cite{zhang2018shufflenet} and MobileNet-v2~\cite{sandler2018mobilenetv2}. Recently, a huge interest in automatically reducing the shape of the deep networks has gained interest, with works on neural network sparsification~\cite{molchanov2017variational, tartaglione2018learning, louizos2017learning, ullrich2017soft} boosted by the recent lottery ticket hypothesis by Frankle~and~Carbin~\cite{frankle2018lottery}. These approaches address the problem of improving inference efficiency with limited memory footprint, but do not directly tackle the problem of reducing stored model size.\\
\paragraph{Minimizing the computation} Recently, this topic is collecting ever-increasing interest. While deepening its roots in statistical physics, some works exploited low-precision training in artificial neural networks~\cite{courbariaux2015binaryconnect, rastegari2016xnor, baldassi2018role}. A large number of works attempts to also use low-precision back-propagation signals and low-precision activations, as it leads to lower power consumption at inference time~\cite{lin2017towards, mishra2017wrpn, zhou2016dorefa}. These techniques, however, do not explicitly address the problem of minimizing the storage size of the entire model.\\
\paragraph{Minimizing the stored model's memory} Here the main goal is not to modify the architecture of a deep model, but to merely compress it, to reduce its stored size: while the other two approaches focused on somehow changing the architecture of the deep model to simplify it and/or to reduce its memory footprint, here the objective is to compress a stored model with no architectural change. Towards this end, many approaches have been proposed: context-adaptive binary arithmetic coding~\cite{wiedemann2019deepcabac}, learning the quantized parameters using the local reparametrization trick~\cite{shayer2017learning}, cluster similar parameters between different layers~\cite{xu2018deep}, using matrix factorization followed by Tucker decomposition~\cite{kim2015compression}, training adversarial neural networks towards compression~\cite{belagiannis2018adversarial} or employing a Huffman encoding scheme~\cite{han2015deep} are just some of them. 
Recently, Oktay~\emph{et~al.} proposed an entropy penalized reparametrizations to the parameters of a deep model, which leads to competitive compression values sacrificing a little bit the deep model's performance~\cite{oktay2019scalable}. However, their approach has some training overhead, like the fact they require to train a decoder, they make their formulation differentiable through the use of straight-through estimators (STE). The big advantage provided by their approach relies more in the re-parametrization leading to the quantization strategy, but the compressibility of their quantized parameters is limited to arithmetic coding.\\
Deep learning based compression schemes proposing a direct high entropy-based regularizer are difficult to design because of the non-differentiability of the entropy and its computational heaviness. All the discussed methods do not explicitly minimize the final compressed file size but they are limited to rigid quantization and compression schemes~\cite{wiedemann2019deepcabac} or they build dictionaries on-purpose~\cite{han2015deep}, losing generality. 
In the next section, we introduce our efficient and differentiable n-th order entropy proxy, to be used in the HEMP framework: it can be freely associated to any quantization strategy and any entropic compression algorithm. Differently to the work by Wiedemann~\emph{et~al.}~\cite{wiedemann2019deepcabac}, HEMP is not bound to a particular quantization scheme, and provides a direct, scalable and differentiable entropy estimator on the continuous parameters.
\section{Entropy-based regularization}
\label{sec:method}

In this section, we describe our entropy-based framework for quantization. We introduce a regularization formulation that
uses a differentiable entropy proxy, evaluated on the continuous parameters of the model, to indirectly reduce the compressed size of the quantized network. We will show that this term easily scales-up to any entropy orders, thus improving the compression efficiency of actual algorithms such as dictionary-based compression.

\subsection{Preliminaries}
\begin{table}
   \center
    \caption{Overview on the notation used in this work.}
    \begin{tabular}{c c}
        \toprule
        \textbf{Symbol} & \textbf{Meaning}\\
        \midrule
        $w_{l,i}$           & $i$-th (continuous) parameter in the $l$-th layer\\
        $\widehat{w}_{l,i}$ & $i$-th (quantized) parameter in the $l$-th layer\\
        $q_{l,i}$           & quantization index corresponding\\
                            & to the $i$-th parameter in the $l$-th layer\\
        $N$                 & quantization levels\\
        $\xi$               & generic quantization index in range $[1; N]$\\
        $p(\widehat{w}_{l,i} \rightarrow \xi)$ & probability that the quantized representation\\
                            & of $w_{l, i}$ will have $\xi$ as quantization index\\
        $\widehat{H}_n$     & $n$-th order entropy on the quantization indices\\
        $H_n$     & differentiable proxy of $\widehat{H}_n$ proposed\\
                    & within HEMP\\
        \bottomrule
    \end{tabular}
    \label{tab:notationover}
\end{table}

Here, we introduce preliminaries and notations. Let a feed-forward, multi-layer artificial neural network be composed of $L$ layers. Let 
$w_{l,i} \in \mathbb{R}$ be the $i$-th parameter of the $l$-th layer.
Let us assume all ANN parameters are quantized onto $N$ discrete levels, with: 
\begin{itemize}[noitemsep,topsep=0pt]
    \item quantization index $q_{l,i}\in [1, N]$ for every parameter $w_{l,i}$;
    \item reconstruction (or representation) levels $r_{l}(k),\ k=1 \ldots N$; as shown in the following, every layer of the ANN model gets its own optimized set of reconstruction levels.
\end{itemize}

From these, we get the quantized parameters $\widehat{w}_{l,i}$ according to
\begin{equation}
    \widehat{w}_{l,i} = r_{l}(q_{l,i}).
\end{equation}
Table~\ref{tab:notationover} collects the most recurring symbols of this section. Please notice that multi-dimensional versions of the symbols are in bold.\\
Now, let us consider $\mathbf{\widehat{w}}^{n}$ as the $n$-uples of the quantized parameters, where $\mathbf{\widehat{w}}_j^{n}$ is the $j$-th $n$-uple of quantized parameters. In general, the $n$-th order entropy on the quantized model is
\begin{equation}
    \label{eq:Hnthquantized}
    \widehat{H}_{n} = -\sum_{\bm{\xi}} p(\mathbf{\widehat{w}}^{n} \rightarrow \bm{\xi}) \log_2 p(\mathbf{\widehat{w}}^{n} \rightarrow \bm{\xi}),
\end{equation}
where $\| W\|_0$ (L0-norm) is the total number of parameters, $\bm{\xi} \in [1;N]^n$ and, using the chain rule, we can express $p(\mathbf{\widehat{w}}^{n} \rightarrow \bm{\xi})$ as
\begin{equation}
    \label{wquantnuple}
    p(\mathbf{\widehat{w}}^{n} \rightarrow \bm{\xi}) = 
    \frac{n}{\| W\|_0} \sum_{j}
    \prod_{m=1}^{n} p\left[ \widehat{w}_{j, m}^n \rightarrow \xi_{m}\left| \bigcap_{s = 1}^{m-1} \right. \left(\widehat{w}_{j, s}^n \rightarrow \xi_{s}\right)\right].
\end{equation}
In \eqref{wquantnuple}, the ``probability'' of the event $\widehat{w}_{j, m}^n \rightarrow \xi_{m}$ is
\begin{equation}
    \label{eq:pbinquantized}
    p(\widehat{w}_{j, m}^n \rightarrow \xi_{m}) = \mathbbm{1}_{\xi_m} \left(q_{l,i}\right)
\end{equation}
where $\mathbbm{1}_{\xi}(\cdot)$ is the indicator function.
Minimizing \eqref{eq:Hnthquantized} results in maximizing the final compression for the quantized model when using an entropic compression algorithm (\cite{witten1987arithmetic,seroussi1993lempel}). 
Unfortunately, the problem in minimizing \eqref{eq:Hnthquantized} within a gradient descent based optimization framework lies in the non-differentiability of \eqref{eq:pbinquantized}. In the next section we introduce a differentiable proxy for \eqref{eq:pbinquantized} which directly optimizes the continuous parameters $w_{l,i}$ such that their quantization is highly compressible.

\subsection{Differentiable n-th order entropy regularization}
\label{sec:1Hcont}
\begin{figure}
    \centering
    \includegraphics[width=0.6\columnwidth]{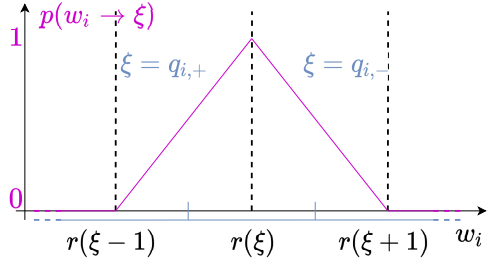}
    \caption{Visual representation of \eqref{eq:pneighbor}.}
    \label{fig::pwi}
\end{figure}

In the previous section we have stated the impossibility of directly optimizing \eqref{eq:Hnthquantized} using gradient descent-based techniques because of the non-differentiability of \eqref{wquantnuple}. We are going to overcome this obstacle providing a formulation of \eqref{wquantnuple} based on the distance between the continuous parameter $w_{l,i}$ and its quantized reconstruction $\widehat{w}_{l,i}$. Here on, we will drop the subscript $l$, but in general all the layers have different reconstruction levels.\\
Let us define first the distance between a parameter $w_{i}$ and the reconstruction level $r(\xi)$:
\begin{equation}
    \label{eq:absdist}
    d[w_{i}, r(\xi)] = \left|w_{i} - r(\xi)\right|
\end{equation}
From~\eqref{eq:absdist}, we can estimate the probability of binning $w_i$ to $\xi$ using the softmax function:
\begin{equation}
    \label{eq:softmaxprob}
    p(w_i \rightarrow \xi) = \frac{e^{-d[w_i, r(\xi)]}}{\sum_{j} e^{-d[w_i, r(j)]}}
\end{equation}
Such general formulation is computationally expensive, so we propose an efficient approximation thereof exploiting a ``bin locality'' principle, for which we say that a parameter $w_i$ can be binned to the two closest bins only. Under the assumption of quasi-static process, indeed, locally the probability of binning the continuous parameter $w_i$ in other bins than the two closest between two iteration steps can be neglected. We refer to these bins as $q_{i,-}$ and $q_{i,+}$. 
In this case, we know $w_i\in [r(q_{i,-}); r(q_{i,+})]$. Here we can design a relative distance linearly-scaling probability:
\begin{equation}
    \label{eq:pneighbor}
    p(w_i \rightarrow \xi)= \left\{
    \begin{array}{cl}
         1 - \frac{w_i - r(q_{i,-})}{\Delta_{i}}  & \xi = q_{i,-}\\
         1 - \frac{r(q_{i,+}) - w_i}{\Delta_{i}}  & \xi = q_{i,+}\\
         0 & otherwise
    \end{array}
    \right .
\end{equation}
where $\Delta_i = r(q_{i,+}) - r(q_{i,-})$. Figure~\ref{fig::pwi} displays the behavior of \eqref{eq:pneighbor}.
Hence, the binning probability in~\eqref{eq:pneighbor} scales as the relative distance from the center of the bin. If we combine \eqref{eq:pneighbor} with \eqref{eq:pbinquantized} and, finally, with \eqref{eq:Hnthquantized}, we obtain
\begin{align}
    \label{eq:Hnth}
    H_{n} = \frac{n}{\| W\|_0} \sum_{\bm{\xi}} & \left\{ \left[\sum_{j}
    p(\mathbf{w}_{j}^n\rightarrow \bm{\xi}) \right] \cdot\right .\nonumber\\
    &\left .\cdot \left[ \log_2(\| W\|_0) - \log_2(n) - \log_2 \sum_{j} p(\mathbf{w}_{j}^n\rightarrow \bm{\xi})\right] \right\}
\end{align}
where
\begin{equation}
    p(\mathbf{w}_{j}^n\rightarrow \bm{\xi}) = \prod_{m=1}^{n} p\left[ w_{j, m}^n \rightarrow \xi_{m}\left| \bigcap_{s = 1}^{m-1} \right. \left(w_{j, s}^n \rightarrow \xi_{s}\right)\right]
\end{equation}

\subsection{Study of the entropy regularization term}

In this section we are detailing the derivations of the proposed entropy regularization. Obtaining an explicit formulation for the update terms allows to efficiently implement the update rule explicitly (when using gradient-based optimizers, without relying on the automatic differentiation packages) and to study both the stationary points of the regularization term and bounds of the gradient respectively.\\
\subsubsection{Explicit derivation of the entropy regularization term's gradient}
\label{sec:H1}
Let us consider here the first order entropy proxy:
\begin{equation}
    \label{eq:H1}
    H_1 = - \sum_{\xi} p(\mathbf{w} \rightarrow \xi) \log_2 \left[ p(\mathbf{w} \rightarrow \xi) \right]
\end{equation}
with
\begin{equation}
    \label{eq:simplep}
    p(\mathbf{w} \rightarrow \xi) = \frac{1}{\|W\|_0} \sum_i p(w_i \rightarrow \xi)
\end{equation}
Let us differentiate \eqref{eq:H1} with respect to $w_i$:
\begin{align}
    &\frac{\partial H_1}{\partial w_i} = - \frac{\partial}{\partial w_i} \left\{\sum_{\xi} p(\mathbf{w} \rightarrow \xi) \log_2 \left[ p(\mathbf{w} \rightarrow \xi) \right]\right\}\nonumber\\
    &= -\sum_{\xi} \left\{ \frac{\partial}{\partial w_i} \left[p(\mathbf{w} \rightarrow \xi)\right] \cdot \log_2 \left[ p(\mathbf{w} \rightarrow \xi) \right] + p(\mathbf{w} \rightarrow \xi) \cdot \frac{\partial}{\partial w_i}  \log_2 \left[ p(\mathbf{w} \rightarrow \xi) \right] \right\}\nonumber\\
    &= -\left\{ \sum_{\xi} \frac{\partial}{\partial w_i} \left[p(\mathbf{w} \rightarrow \xi)\right] \cdot \log_2 \left[ p(\mathbf{w} \rightarrow \xi) \right] +\right .\nonumber\\
    &~~~~~~~~~~~~~~~~~~~~~~~~~~~~~~~~~~~~~~~~~~~~~~~~~~~~~~~~\left .+\sum_{\xi} p(\mathbf{w} \rightarrow \xi) \cdot \frac{\partial}{\partial w_i}  \log_2 \left[ p(\mathbf{w} \rightarrow \xi) \right] \right\}
    .
\end{align}
According to \eqref{eq:pneighbor}, we can write
\begin{align}
     \frac{\partial H_1}{\partial w_i} = - &\left\{ \sum_{\xi} \frac{\partial}{\partial w_i} \left[p(\mathbf{w} \rightarrow \xi)\right] \cdot \log_2 \left[ p(\mathbf{w} \rightarrow \xi) \right] +\right .\nonumber\\
      +&  \sum_{\xi=\{q_{i, -}, q_{i,+}\}} \left . p(\mathbf{w} \rightarrow \xi) \cdot \frac{\partial}{\partial w_i}  \log_2 \left[ p(\mathbf{w} \rightarrow \xi) \right] \right\} , \label{eq:tempH1der} 
\end{align}
and considering that
\begin{equation}
    \frac{\partial}{\partial w_i} p(\mathbf{w} \rightarrow \xi)= \left\{
    \begin{array}{ccl}
         -&\frac{1}{\|W\|_0 \Delta_{i}}  & \xi = q_{i,-}\\\\
         &\frac{1}{\|W\|_0 \Delta_{i}}  & \xi = q_{i,+}\\\\
         &0 & otherwise
    \end{array}
    \right .
\end{equation}
where 
\begin{equation}
    \Delta_i = r(q_{i,+}) - r(q_{i,-}),
\end{equation}
we have
\begin{align}
     \frac{\partial H_1}{\partial w_i} &= - \sum_{\xi=\{q_{i, -}, q_{i,+}\}} \left\{ \frac{\partial}{\partial w_i} \left[p(\mathbf{w} \rightarrow \xi)\right] \cdot \log_2 \left[ p(\mathbf{w} \rightarrow \xi) \right] \right\}\nonumber\\
     &= \frac{1}{\|W\|_0 \Delta_{i}} \log_2 \frac{p(\mathbf{w} \rightarrow q_{i,-})}{p(\mathbf{w} \rightarrow q_{i,+})}~.
     \label{eq:H1der}
\end{align}

\noindent Using a similar approach, we can explicitly write the gradient for the n-th order entropy term in \eqref{eq:Hnth}:
\begin{equation}
    \label{eq:Hderivativenth}
    \frac{\partial H_n}{\partial w_{j, m}^n} = \frac{n}{\Delta_{j, m}^n \| W\|_0} \log_2\left\{ \frac{\displaystyle
    \prod_{\{\bm{\xi}\}_j,~\xi_m = \xi_{j, m,-}^n}  p(\mathbf{w}_{j}^n\rightarrow \bm{\xi})
    }{\displaystyle
    \prod_{\{\bm{\xi}\}_j,~\xi_m = \xi_{j, m,+}^n}  p(\mathbf{w}_{j}^n\rightarrow \bm{\xi})
    } 
    \right\}
\end{equation}
where $\{\bm{\xi}\}_j$ indicates the set of $\bm{\xi}$ whose binning probability for $\mathbf{w}_{j}^n$ is non-zero.\\
Having \eqref{eq:Hderivativenth} explicit enables efficient gradient computation: indeed, given the designer choice in \eqref{eq:pneighbor}, every $n$-uple of parameters has $2^n$ possible quantization indices $n$-uples
$\{\bm{\xi}\}$ only, which is independent on the number of quantization levels $N$. On the contrary, using \eqref{eq:softmaxprob} would result in $N^n$ possible quantization indices $n$-uples $\forall\ \mathbf{w}_j^n$. Hence, out proposed approach allows us to save $(N/2)^n \times $ memory at computation time.\\
For sake of simplicity, the following analysis on stationary points and boundaries will be performed on the first order entropy, but similar conclusions can be equivalently drawn for any n-th order.

\subsubsection{Stationary points for H1}
\label{sec:spH1}
\begin{figure}
    \centering
    \includegraphics[width=1.0\columnwidth]{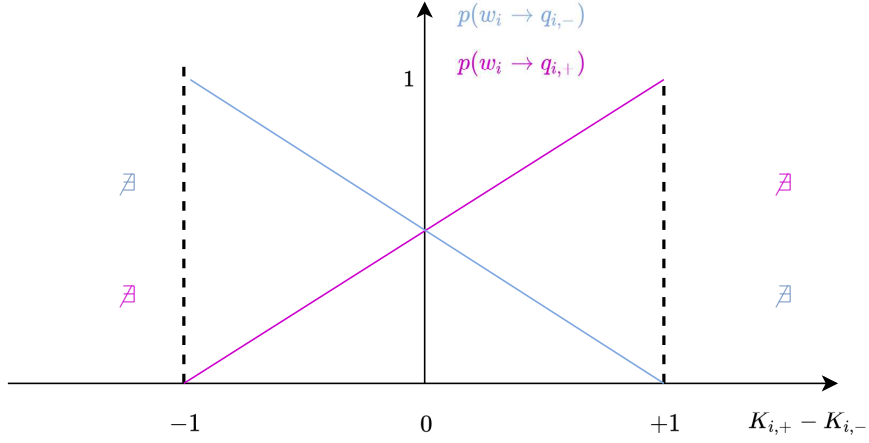}
    \caption{Gradient vanishing condition for both $ p(w_i~\rightarrow~q_{i,-})$ (in cyan) and $p(w_i~\rightarrow~q_{i,+})$ (in violet).}
    \label{fig::zerocond}
\end{figure}%

In this section we are looking for $H_1$ stationary points (or in other words, when gradient vanishes). From \eqref{eq:H1der} we observe that
\begin{equation}
    \label{eq:zeroimplicit}
    \frac{\partial H_1}{\partial w_i}=0 \Leftrightarrow p(\mathbf{w} \rightarrow q_{i,-}) = p(\mathbf{w} \rightarrow q_{i,+}),
\end{equation}
assuming $\Delta_i, \|W\|_0$ finite positive numbers. We can make $w_i$ explicit in the condition \eqref{eq:zeroimplicit}:
\begin{equation}
    \label{eq:zerocond}
    \left[ p(w_i \rightarrow q_{i,+}) - p(w_i \rightarrow q_{i,-}) \right] = K_{i,+} - K_{i,-}
\end{equation}
where
\begin{align*}
    K_{i,+} &= \sum_{j \neq i}{p(w_j\rightarrow q_{i,+})}\\
    K_{i,-} &= \sum_{j \neq i}{p(w_j\rightarrow q_{i,-})}~.
\end{align*}
According to \eqref{eq:pneighbor}, we can rewrite \eqref{eq:zerocond} as
\begin{equation}
    \label{eq:zerocond2}
    p(w_i \rightarrow q_{i,+}) = \left\{
    \begin{array}{l l}
        \displaystyle \frac{1}{2}\left(K_{i,+} - K_{i,-} + 1\right)& if~(K_{i,+} - K_{i,-})\in\left[-1;+1\right]\\\\
        \displaystyle\nexists &otherwise
    \end{array}
    \right .
\end{equation}
because $p(w_i \rightarrow q_{i,+})\in [0;1]$ by definition. As we expect, if $q_{i,+}$ and $q_{i,-}$ are evenly populated, $K_{i,+}=K_{i,-}$ and the stationary point of $\frac{\partial H_1}{\partial w_i}$ is
\begin{equation}
    p(w_i  \rightarrow q_{i,+}) = p(w_i \rightarrow q_{i,-}) = \frac{1}{2}\nonumber
\end{equation}
which results in
\begin{equation}
     w_i = \frac{1}{2}\left[r(q_{i,+}) + r(q_{i,-})\right],
\end{equation}
exactly between the centre of the two bins. From the entropy point of view, this is essentially what we expect, since we have two equi-populated bins; however, this is not what we like to have when we quantize a deep network, considering that it leads to an high quantization error. For this reason, favoring solutions in which $p(w_i \rightarrow \xi)\neq \frac{1}{2}$ is a good strategy and this is also the reason we included a reconstruction error in the overall regularization function.

\subsubsection{Bound for H1's derivative}
\label{sec:H1bound}
\begin{figure}
    \centering
    \includegraphics[width=0.8\columnwidth]{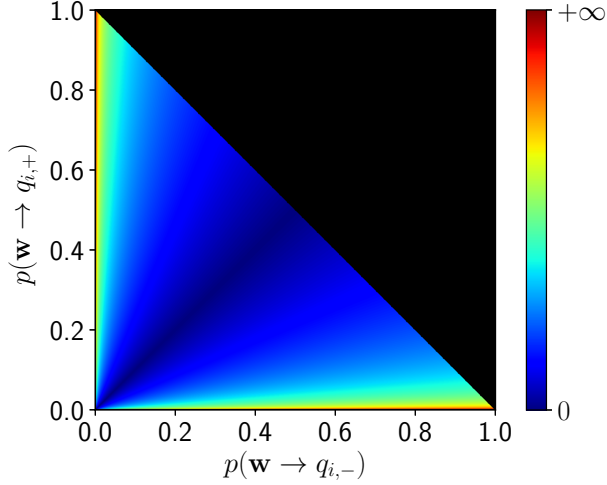}
    \caption{Plot of the absolute upper bound \eqref{eq:bound} for $\frac{\partial H_1}{\partial w_i}$ as a function of $p(\mathbf{w}~\rightarrow~q_{i,-})$ and $p(\mathbf{w}~\rightarrow~q_{i,+})$. In black we represent regions out of the considered domain.} 
    \label{fig::derivbound}
\end{figure}%
In this section we are looking for an upper bound of $\frac{\partial H_1}{\partial w_i}$ and we study the cases in which such quantity explodes, in order to assess conditions to avoid gradient explosion. We can set the bound for the gradient magnitude as:
\begin{equation}
    \label{eq:bound}
    \frac{\partial H}{\partial w_i} \leq  \left| \frac{1}{\Delta_i \| W\|_0 } \log_2 \frac{p(\mathbf{w} \rightarrow q_{i,-})}{p(\mathbf{w} \rightarrow q_{i,+})}\right|~.
\end{equation}
Considering that $\| W\|_0\in \mathbb{N}$ and that $\Delta_i> 0$ (so, both are finite, real-valued quantities), we are interested to guarantee
\begin{equation}
    \frac{\partial H}{\partial w_i} \leq  \frac{1}{\Delta_i \| W\|_0 } \left| \log_2 \frac{p(\mathbf{w} \rightarrow q_{i,-})}{p(\mathbf{w} \rightarrow q_{i,+})}\right| < K
\end{equation}
where $K$ is a positive real-value finite number. Given $p(\mathbf{w} \rightarrow \xi) \in [0;1]$, let us study the cases in which such quantity explodes.
\begin{itemize}
    \item Case $\frac{p(\mathbf{w} \rightarrow q_{i,-})}{p(\mathbf{w} \rightarrow q_{i,+})} \rightarrow 0^+$. In this case we have $p(\mathbf{w} \rightarrow q_{i,-})\rightarrow 0^+$ and $p(\mathbf{w} \rightarrow q_{i,+})\neq 0$. According to \eqref{eq:H1der} $w_i\in [r(q_{i,-}); r(q_{i,+}))$; so at least one parameter lies in the considered interval and the condition is impossible by construction.
    \item Case $\frac{p(\mathbf{w} \rightarrow q_{i,-})}{p(\mathbf{w} \rightarrow q_{i,+})} \rightarrow +\infty$. In this case we have $p(\mathbf{w} \rightarrow q_{i,+})\rightarrow 0^+$ and $p(\mathbf{w} \rightarrow q_{i,-})\neq 0$. Similarly to the previous case, $w_i\in [r(q_{i,-}); r(q_{i,+}))$; so at least one parameter lies in the considered interval and the condition is impossible by construction.
    \item Case $\frac{p(\mathbf{w} \rightarrow q_{i,-})}{p(\mathbf{w} \rightarrow q_{i,+})} \rightarrow \frac{0^+}{0^+}$. By construction, $w_i\in [r(q_{i,-}); r(q_{i,+}))$ so this case is impossible.
\end{itemize}
In the next section we will describe the overall HEMP framework.

\section{Training scheme}
\label{sec:steps}


\begin{figure*}[t]
    \centering
    \includegraphics[width=1.0\textwidth]{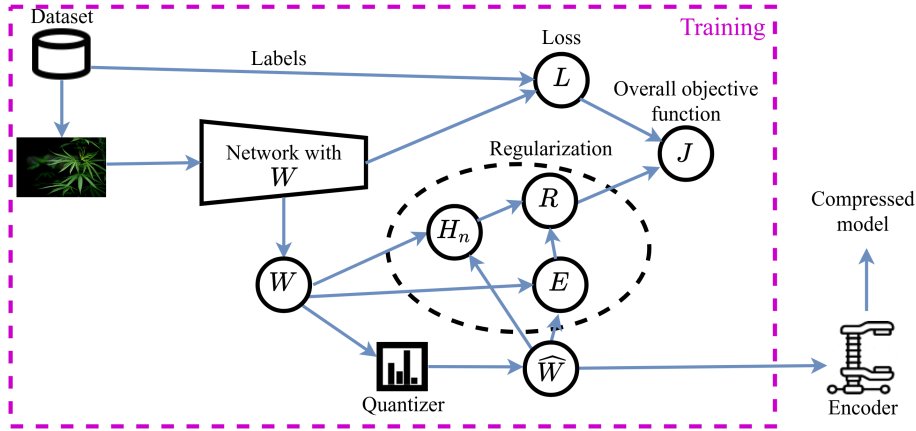}
    \caption{Schematic representation of HEMP.}
    \label{fig::generalscheme}
\end{figure*}%




The overall training scheme is summarized in Fig.~\ref{fig::generalscheme} and includes a quantizer and an entropy encoder. The quantizer generates the discrete-valued representation $\widehat{W}$ of the network parameters at training time. The encoder produces the final compressed file embedding the deep model once the training is over. 
Our scheme does not make any assumption about the quantization or entropy coding scheme,
contrarily to other strategies tailored for, e.g., specific quantization schemes (\cite{han2015deep, wiedemann2019deepcabac}).
Therefore, in the following we will assume a very general, non-uniform Lloyd-max quantizer, 
while we will not make any assumption about the entropy encoder for the moment as it is external to the training process.
Our learning problem can be formulated as follow: given a dataset and a network architecture, we want to compress the network parameters $W$, while preserving the network performance as measured by some loss function $L$.
Towards this end, we introduce the following regularization function:
\begin{equation}
    R = \lambda_H H_n + \lambda_E E
    \label{eq:RHE}
\end{equation}
\noindent
where $\lambda_H$ and $\lambda_E$ are two positive hyper-parameters and 

\begin{equation}
    \label{eq:E}
    E = \sqrt{\frac{1}{\|W\|_0} \sum_l \sum_i \left[w_{l,i} - r_l(q_{l,i})\right]^2}
\end{equation}
\noindent
is a reconstruction error estimator. Minimizing $E$ makes $w_{l,i} \rightarrow \widehat{w}_{l,i}$ and, for instance, loss evaluation on the continuous parameters network approaches the loss estimated on the quantized network.
Overall, we minimize the objective function:
\begin{equation}
    \label{eq:J}
    J = L + R
\end{equation}
Minimizing $J$ requires finding the right balance between $L$ and $R$: towards this end, we propose to dynamically re-weight $R$ according insensitivity of each parameter (\cite{tartaglione2018learning}). The key idea here is to re-weight the regularization gradient $\frac{\partial R}{\partial w_{l,i}}$ at every parameter update depending on the sensitivity of the loss $L$ with respect to every parameter. We say that the larger the magnitude of the gradient of the loss with respect to $w_i$, the smaller the perturbation from the minimization induced by $R$ we desire. Hence, in the update of the parameter $w_{l,i}$, we re-weight the gradient of $R$ by the insensitivity:
\begin{equation}
    \label{eq:insensitivity}
    \bar{S}_{l,i} = 1-\frac{\left|\frac{\partial L}{\partial w_{l,i}}\right|}{\max_j \left\{\left|\frac{\partial L}{\partial w_{l,j}}\right|\right\} }
\end{equation}
The HEMP framework allows to solve the learning problem using standard optimization strategies, where the gradient of \eqref{eq:J} is descended.

\section{Experiments}
\label{sec:experiments}
In this section we evaluate the effectiveness of HEMP. Towards this end, we propose experiments on several widely used datasets with different architectures.
\paragraph{Datasets and architectures}
We experiment with LeNet-5 on MNIST, ResNet-32 and MobileNet-v2 on CIFAR-10, ResNet-18 and ResNet-50 on ImageNet. We always train from scratch except for ImageNet experiments where we rely on pre-trained models.\footnote{https://pytorch.org/docs/stable/torchvision/models.html}
\paragraph{Setup} We experiment on a Nvidia~RTX~2080~Ti GPU. Our algorithm is implemented using PyTorch~1.5.\footnote{The source code will be made available on GitHub upon acceptance of the work.}
For all our simulations we use SGD optimization with momentum $0.9$, $\lambda_{H}=1$ and $\lambda_{E}=0.1$. Learning rate and batch-size depend on the dataset and the architecture: for all the datasets except for ImageNet the learning rate used is $10^{-2}$ and batch-size $100$, for ResNet-18 trained on ImageNet the learning rate is $10^{-3}$ with batch-size $128$ while for ResNet-50 learning rate is $10^{-4}$ with batch-size 32.
The file containing the quantized parameters is entropy-coded using LZMA~\cite{pavlov2007lzma}, a popular dictionary-based compression algorithm well-suited to exploit high-order entropy.
\paragraph{Metrics} The goal of the present work is to compress a neural network without jeopardizing its accuracy, so 
we rely on two distinct, largely used, performance metrics:
\begin{itemize}
    \item the compressed model size as the size of the file containing the entropy-encoded network,
    \item the classification accuracy of the compressed network (indicated as Top-1 in the following).
\end{itemize}
\subsection{Preliminary experiments}

As preliminary experiments, we evaluate if the regularizer function \eqref{eq:Hnth} is a good estimator of \eqref{eq:Hnthquantized}.
Towards this end, we train the LeNet-5 architecture on MNIST minimizing $H_n$ while logging the entropy on the quantized parameters $\widehat{H}_n$. 
\begin{figure*}
    \subfloat[LeNet-5 trained on MNIST]{\scalebox{.48}{\input{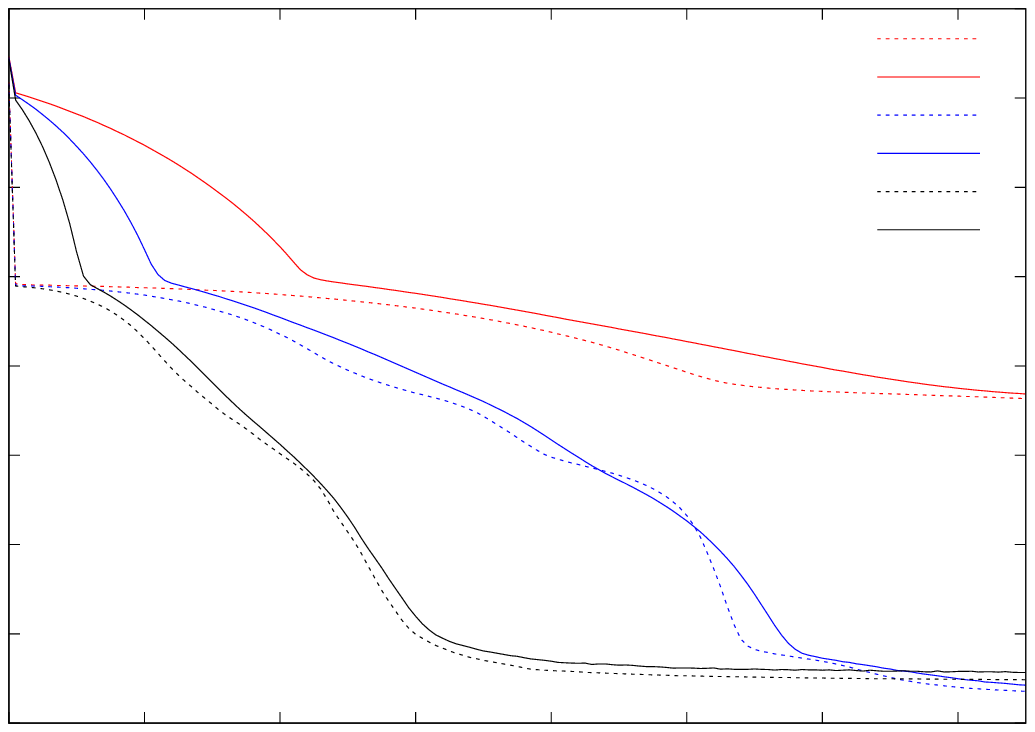}}\label{fig::variousH}}
    \hfill
    \subfloat[ResNet-32 trained on CIFAR-10]{\scalebox{.48}{\input{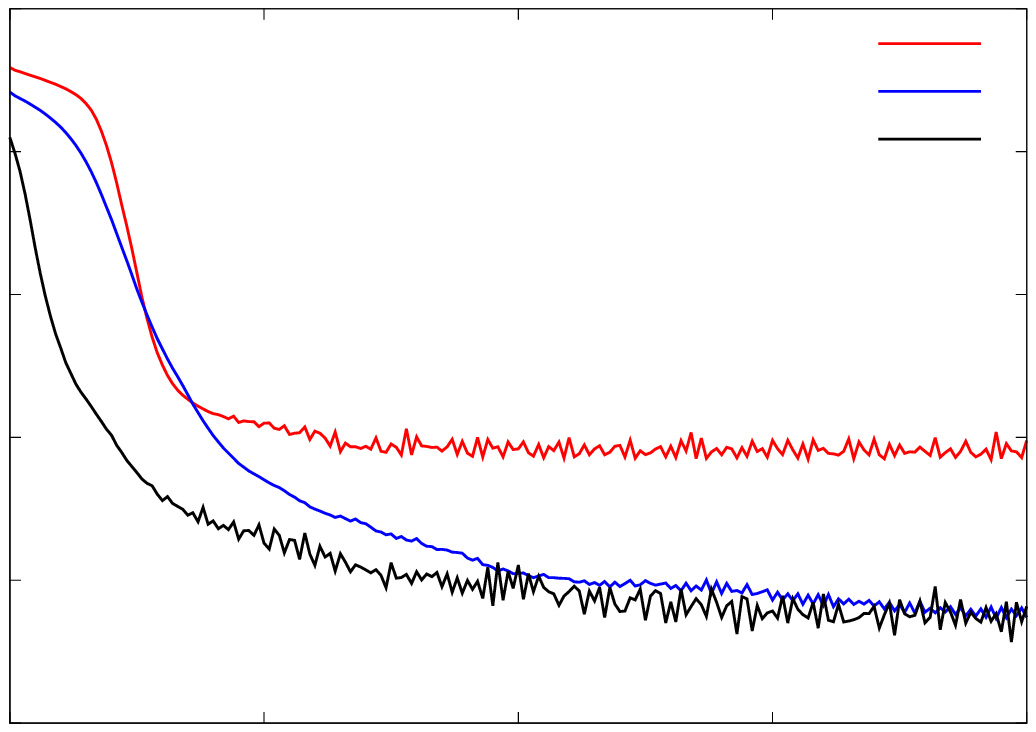}}\label{fig::variousH-2}}
    \hspace{0.5cm}
    \caption{Different entropy order minimization for LeNet-5 trained on MNIST (a) and on ResNet-32 trained on CIFAR-10 (b): in red first order is minimized, in blue the second one and in black the fourth~(a)~/~third~(b). Continuous lines represent the differentiable quantity introduced in \eqref{eq:Hnth} while dashed lines are the actual entropies directly computed on the quantized architecture \eqref{eq:Hnthquantized}.}
\end{figure*}
Fig.~\ref{fig::variousH} shows the normalized $\widehat{H}_n$ and its approximation $H_n$: three findings are noteworthy.\\
\indent First, $H_n$ accurately estimates $\widehat{H}_n$, i.e. minimizing $H_n$ yields to minimizing $\widehat{H}_n$. Under the assumption the quantized parameters are entropy-coded, minimizing $\widehat{H}_n$ shall minimize the size of a file where the encoded parameters are stored.\\
\indent Second, when $n=1$, the training converges to a higher entropy, while minimizing higher entropy orders enables access to lower entropy embeddings. Higher entropy reflects on the final size of the model: while for $n=1$ we could get a final network size of 61kB, for $n=\{2,~4\}$ the final size drops to approximately 27.5kB, having a top-1 accuracy of 99.27\%. This better performance can be explained by the fact that  higher order entropy can catch repeated sequences of parameters' binnings which can lead to a significant compression boost.\\
\indent Third, the higher $n$, the fewer the epochs required to converge to low entropy values. However, in terms of actual training time, the available GPU memory limits the parallelism degree for computing the derivative term in \eqref{eq:Hderivativenth}. In the following, we will stick to $n=2$ as it enables both reasonably low entropy embeddings and training times.\\
\indent As a further verification, we have run the same experiments on the ResNet-32 architecture trained on the CIFAR-10 dataset: also here, we minimize $H_n$ while logging the entropy on the quantized parameters $\widehat{H}_n$ at different values of $n$. Fig.~\ref{fig::variousH-2} shows the normalized $\widehat{H}_n$. Similarly to what observed in the main paper, second-order entropy minimization results to be a good trade-off between complexity and final performance, considering that the reached entropic rate of $n=2$ is comparable to $n=3$. Please notice also that the entropy estimated on the quantized model, and reported Fig.~\ref{fig::variousH-2}, is proportional to the final file sizes.\\
\begin{figure}
    \center
    \includegraphics[width=0.7\textwidth]{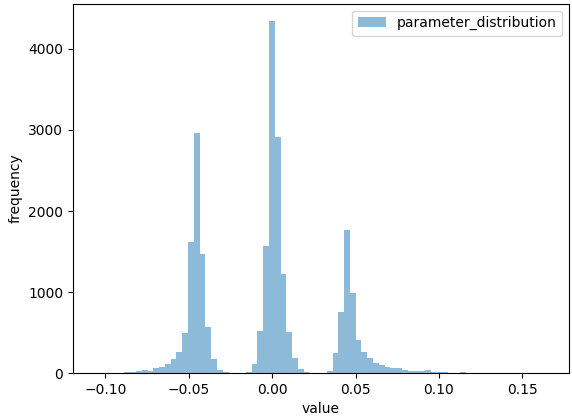}
    \caption{Typical distribution of $w$ values during HEMP optimization for LeNet-5 trained on MNIST (second convolutional layer), with $N=3$.}
    \label{fig:HEMPdistr}
\end{figure}
As a further analysis of HEMP's effect of the parameter, in Fig.~\ref{fig:HEMPdistr} we show the distribution of the optimized parameters on the second convolutional layer in LeNet-5 trained on MNIST (the other layers follow a similar distribution). In this case we optimize the model having 3 quantized values. As we observe the continuous $w$ values are distributed tightly around their quantized representations $\hat{w}$: as $w\rightarrow \hat{w}, L(w)\rightarrow L(\hat{w})$; in the end the accuracy of the quantized representation of the model approaches the accuracy of the continuous model. Additionally, as observed in Fig.~\ref{fig::variousH}, also the entropy of the quantized model is minimized, achieving both a quantized trained model with high accuracy and high compressibility of its representation.

\vspace{0.2cm}
\subsection{Comparison with the state-of-the-art}
\label{sec:compsota}
\begin{table}
   \center
    \caption{Results on the MNIST dataset using LeNet-5 architecture.}
    \label{tab:results-MNIST}
    \begin{tabular}{c c c c c }
        \toprule
        \textbf{Model}              &\textbf{Method}                         & \textbf{Top-1 [\%]} & \textbf{Size} \\
        \midrule
   \multirow{7}{*}{LeNet-5}            &Baseline                   &99.30      &1.7MB\\
      &LOBSTER~\cite{preprint:lobster} &99.10      &19kB\\
      &Han~\emph{et al.}~\cite{han2015deep}&99.26      &44kB\\
      &Wiedemann \emph{et al.}~\cite{wiedemann2019deepcabac} &99.12      &43.4kB\\
      &HEMP                           &99.27      &27.5kB\\
      &Wiedemann \emph{et al.}(+pruning)~\cite{wiedemann2019deepcabac} &99.02      &11.9kB\\
      &HEMP+LOBSTER~~\cite{preprint:lobster}  &99.05      &2.00kB\\
        \bottomrule
    \end{tabular}
\end{table}

\begin{table}
   \center
    \caption{Results on CIFAR-10 using different architectures.}
    \label{tab:results-CIFAR}
    \begin{tabular}{ c c c c }
        \toprule
        \textbf{Model}              &\textbf{Method}                         & \textbf{Top-1 [\%]} & \textbf{Size} \\
        \midrule
           \multirow{4}{*}{ResNet-32}          
              &Baseline                   & 93.10     &1.9MB \\
              &LOBSTER~\cite{preprint:lobster} & 92.97     &439.4kB\\
              &HEMP                           & 91.57     &168.3kB \\
              &HEMP+ LOBSTER ~\cite{preprint:lobster}& 92.55     &86.2kB \\
        \midrule
                    \multirow{2}{*}{MobileNet-v2}       
                                        &Baseline                   & 93.67     &9.4MB \\
                                       &HEMP                           & 92.80     &872kB \\
                \bottomrule
    \end{tabular}
\end{table}
\begin{table}
   \center
   \caption{Results on ImageNet using different architectures.}
    \label{tab:results-imagenet}
     \begin{tabular}{ c c c c }
      \toprule
        \textbf{Model}              &\textbf{Method}                         & \textbf{Top-1 [\%]} & \textbf{Size} \\
        \midrule
      \multirow{6}{*}{ResNet-18}          
                &Baseline                   & 69.76       &46.8MB \\
                &LOBSTER~\cite{preprint:lobster}&70.12          &17.2MB\\
                &Lin~\emph{et al.}~\cite{lin2017towards}         &68.30     &5.6MB\\
                &Shayer~\emph{et al.}~\cite{shayer2017learning}     &63.50     &2.9MB\\
                &HEMP                           &68.80      &3.6MB  \\
                &HEMP+LOBSTER~\cite{preprint:lobster}&69.70      &2.5MB  \\
        \midrule

                    \multirow{7}{*}{ResNet-50}          &Baseline                   &76.13   &102.5MB \\
                                &Wang \emph{et al.} \cite{wang2019haq}            &70.63      &6.3MB\\
                                &Han \emph{et al.}\cite{han2015deep}            &68.95      &6.3MB\\
                                &Wiedemann \emph{et al.} \cite{wiedemann2019deepcabac} &74.51      &10.4MB\\
                                &Tung \emph{et al.} \cite{tung2018deep}  &73.7               &6.7MB\\
                                &HEMP (high acc.)              &74.52      &9.1MB  \\
                                &HEMP                           &71.33      &5.5MB  \\
        \midrule
                    \multirow{5}{*}{MobileNet-v2}         &Baseline                   &72.1  &13.5MB \\
                                                                    &Tu~\emph{et~al.}\cite{tu2020pruning}   & 7.25    &10.1MB\\
                                                                    &He~\emph{et~al.}\cite{he2019real}      &9.8       &4.95MB\\
                                                                    &Tung \emph{et al.}\cite{tung2018deep}  &70.3               &2.2MB\\
                                &HEMP                       &71.3      &1.7MB  \\
        \midrule
                    Custom, latency 6.11ms,          &APQ~\cite{Wang2020APQ}     &72.8  &20.8MB \\
                    energy 9.14mJ            &APQ~\cite{Wang2020APQ} + HEMP &72.5      &3.04MB  \\
        \bottomrule
    \end{tabular}
\end{table}

\paragraph{HEMP} We now compare our method with state-of-the-art methods for network compression. Our main goal is to minimize the size of the final compressed file while keeping the top-1 performance as close as possible to the baseline network's one. Therefore, our approach can be compared only with works that report the real final file size. To the best of our knowledge, only the methods reported in Tables~\ref{tab:results-MNIST},~\ref{tab:results-CIFAR}~and~\ref{tab:results-imagenet} can be included in this compression benchmark. Indeed, most of the pruning-based methods (\cite{molchanov2017variational, tartaglione2018learning}) typically report pruning-rates only, which can not be directly mapped to file size: encoding sparse structures requires additional memory to store the coordinates for the un-pruned parameters. We implemented one state-of-the-art pruning method (LOBSTER~\cite{preprint:lobster}) to report pruning baseline storage memory achieved. Concerning quantization methods, existing approaches either focus on quantization to boost inference computation minimization (\cite{courbariaux2015binaryconnect, rastegari2016xnor, lin2017towards, mishra2017wrpn, zhou2016dorefa}) or do not report the final file size (\cite{ullrich2017soft, kim2015compression, belagiannis2018adversarial, xu2018deep}). 
We also tried to directly compress the baseline file and we did not observe any compression gain. Therefore, to make reading easier, we did not report these numbers.\\
\indent As a first experiment, we train LeNet-5 on MNIST (Table~\ref{tab:results-MNIST}): despite the simplicity of the task, the reference LeNet-5 is notoriously over-parametrized for the learning task. Indeed, as expected, most of the state-of-the-art techniques are able to compress the model to approximately 40kB. In such a context, HEMP performs best, lowering the size of the compressed model to 27.5kB.\\
\indent Then, we experiment with ResNet-32 and MobileNet-v2 on CIFAR-10 (as reported in Table~\ref{tab:results-CIFAR}), achieving also in this case significant compression: ResNet-32 size drops from 1.9MB to 168kB and MobileNet-v2 from 9.4MB to 822kB. Note that, other literature methods do not report experiments on CIFAR-10 on the proposed architectures. Nevertheless, HEMP approximately reduces the network size by a factor 11 for both architectures.\\
\indent We also compress pretrained ResNet-18, ResNet-50 and MobileNet-v2 trained on ImageNet (Table~\ref{tab:results-imagenet}). Also in this case, HEMP reaches competitive final file size, being able to compress ResNet-18 from 46.8MB to 3.6MB with minimal performance loss and ResNet-50 from 102.5MB to 5.5MB. For the ResNet-50 experiment, we also report partial result for high accuracy band, indicated as ``high acc'', to compare to Wiedemann~\emph{et al.}~\cite{wiedemann2019deepcabac}: for the same accuracy, HEMP proves to drive the model to a higher compression. In the case of ResNet-18, \cite{shayer2017learning} achieves a 0.5MB smaller compressed model, which is however set off by a 4.3\% worse Top-1 error. Also in the case of very efficient architectures like Mobilenet-v2, HEMP is able to reduce significantly the storage memory occupation, moving from 13.5MB to 1.7MB only. Furthermore, the error drop is in this case very limited (0.8\%) when compared to other techniques, like Tung~\emph{et al.} which, in the case of less optimized architectures like ResNet-50, do not have a large drop. While concurrent techniques rely on typical pruning+quantization strategies, aiming at \emph{indirectly} eliminating the redundancy in the models, HEMP is \emph{directly} optimizing over the existing redundancy.\\
Finally, we also tried to make HEMP cope with a different quantize and prune scheme. In particular, APQ~\cite{Wang2020APQ} proves to be a perfect framework for our purpose, since it is a strategy performing both network architecture search, pruning and quantization. We have used HEMP in the most challenging scenario proposed by Wang~\emph{et~al.}, with the lowest latency constraint (6.11ms) and the lowest energy consumption (9.14mJ) at inference time. In this case, we have fine-tuned the APQ's provided model for 5 epochs. Even in this case, HEMP is able to reduce the model's size, from the 20.8MB of the model to 3.04MB only, proving on-the-field its deployability as a companion besides other quantization/pruning scheme, and non-exploiting any prior on the network's architecture.\\
\indent Overall, these experiments show that HEMP strikes a competitive trade-off between compression ratio and performance on several architectures and datasets.
\paragraph{HEMP+LOBSTER} It has been observed that combining pruning and compression techniques enhances reduces the model final file size with little performance loss~\cite{wiedemann2019deepcabac}. In our context, this translates into including two constraints to the learning:
\begin{itemize}
    \item force the quantizer to have, for some $\xi$, the representation $r(\xi)=0$ (or in simpler words, a quantization level corresponding to ``0'');
    \item include a pruning mechanism (permanent parameter set to ``0''). 
\end{itemize}
Both of the constraints work independently from HEMP: indeed, HEMP is not a quantization technique, but it is thought to side any other learning strategy whose aim is to quantize the model's parameters (in such context, pruning ``quantizes to zero'' as many parameters as possible). Hence, we tried to side HEMP to LOBSTER~\cite{preprint:lobster}, which is a state-of-the-art differentiable pruning strategy (hence, compatible within HEMP's framework).\\
\indent The results are as well reported in Tables~\ref{tab:results-MNIST},~\ref{tab:results-CIFAR}~and~\ref{tab:results-imagenet}: it is evident that, including a prior on the optimal distribution of the parameters (removing all the un-necessary ones for the learning problem) helps HEMP to compress more. We have tested the setup HEMP~+~LOBSTER on one architecture per dataset: LeNet-5 (MNIST), ResNet32 (CIFAR-10) and ResNet-18 (ImageNet). While LOBSTER alone is able to achieve highly compressed models for toy datasets (like MNIST), it can not achieve high compression alone on more complex datasets. Still, siding a technique like LOBSTER to HEMP, boosts the compression of 10x for MNIST and ImageNet dataset and 4x for CIFAR-10.\\
\indent HEMP minimizes the $n$-th entropy order (in these experiments, $n=2$) - or in other words, maximizes the occurrence of certain sequences of quantization indices. The mapping of these quantization indices to quantization levels has to be determined outside HEMP: when we run experiments with ``HEMP'' alone, the loss minimization (in our case, the cross-entropy) automatically determines these levels - with the general-purpose Lloyd-max quantizer. However, pruning strategies include a prior on one of the quantization levels (the one corresponding to ``0''), and this helps towards having a higher entropy minimization.

\subsection{Ablation study}
\label{sec:ablation}
\begin{figure*}
    \subfloat[Reconstruction error regularization]{\scalebox{.5}{\input{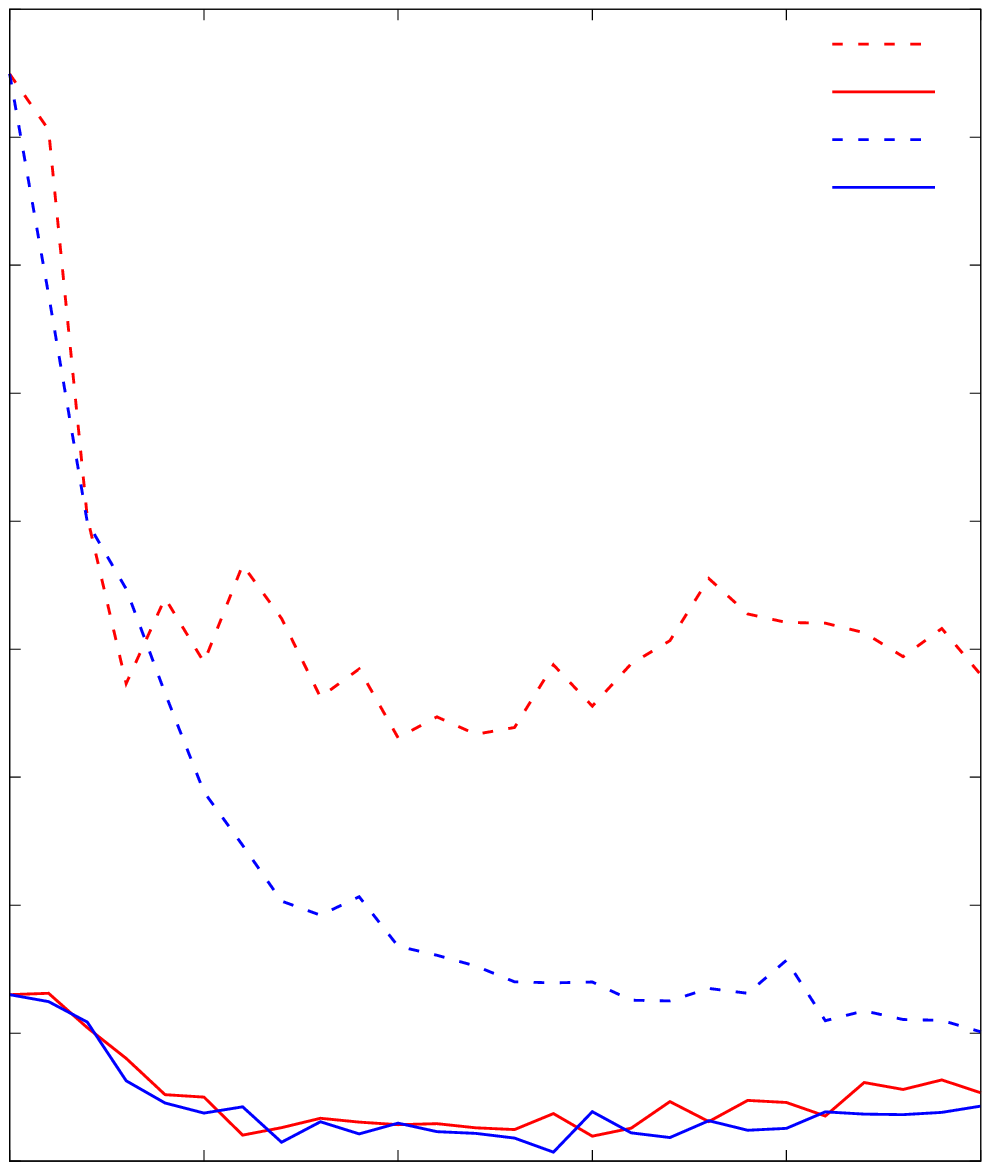}}\label{fig::noE}}
    \hfill
    \subfloat[Insensitivity re-weighting]{\scalebox{.5}{\input{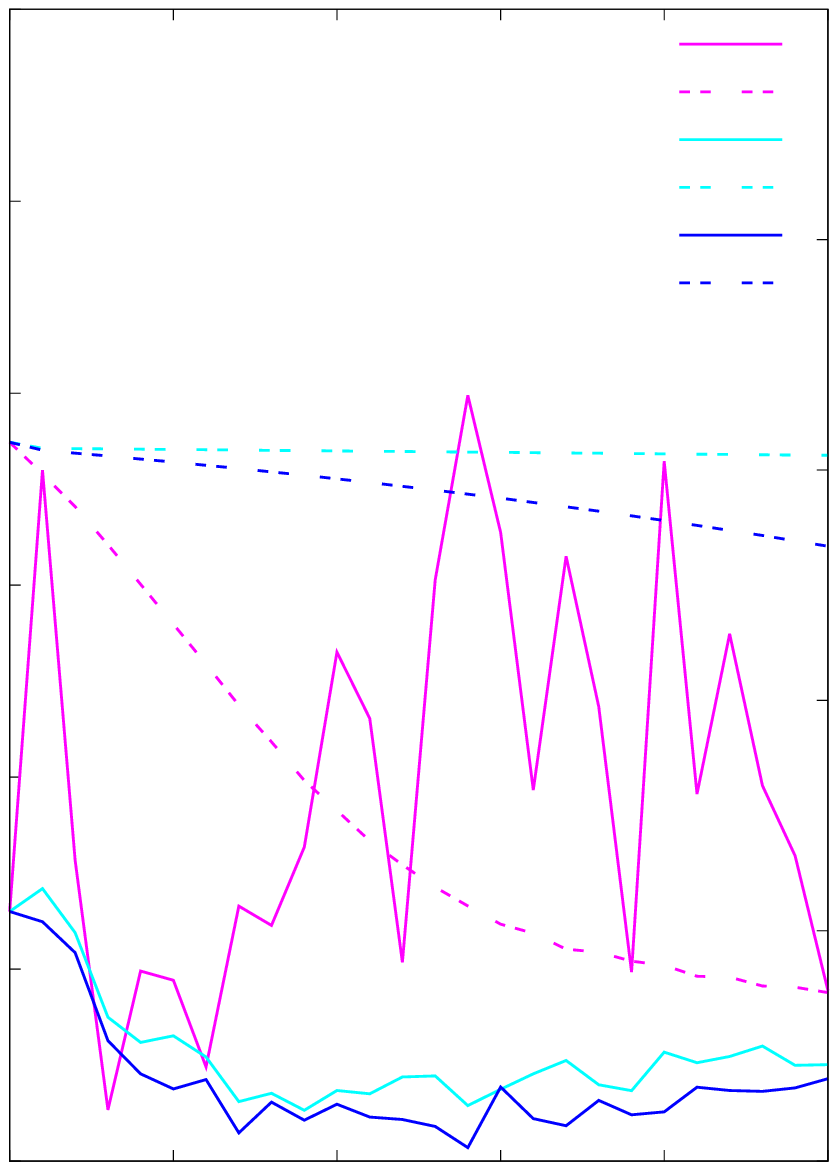}}\label{fig::noS}}
    \caption{Test set losses for different trainings on ResNet-32 trained on CIFAR-10 (a), and effect of the insensitivity as re-weighting factor for $R$ (b). Please notice that the blue line in both (a) and (b) refers to the same simulation, which refers to the standard HEMP training.}
    \label{fig::noES}
\end{figure*}
Here, we evaluate the impact of the reconstruction error term \eqref{eq:E} and the overall insensitivity re-weighting \eqref{eq:insensitivity} for the regularization function. Towards this end, we perform an ablation study on the ResNet-32 architecture trained on CIFAR-10.
\paragraph{Reconstruction error regularization} Fig.~\ref{fig::noE} (left), shows the ResNet-32 loss for the continuous and the quantized models ($L(W)$ and $L(\widehat{W})$ when the reconstruction error $E$ is included or excluded ($\lambda_E=0$) from the regularization function \eqref{eq:RHE}.
We observe that both continuous models (solid lines) obtain similar performance on the test set. However, the quantized models (dashed lines) perform very differently. When the reconstruction error is not included in the training procedure (red lines), the quantized model reach a plateau with a high loss value showing that the network performs poorly on the test set. Conversely, when the reconstruction error is included (blue lines), the quantized model reaches a final loss closer to the continuous models. Indeed, regularizing also on \eqref{eq:E} makes $w_{l,i}\rightarrow \hat{w}_{l,i}~\forall~l,i$, hence $L(W)\rightarrow L(\widehat{W})$. This experiment verifies the contribution of the error reconstruction regularization term towards the good performance of the quantized model.
\paragraph{Insensitivity-based re-weighting} Fig.~\ref{fig::noS} (right) shows the performance of the ResNet-32 model including or excluding the insensitivity re-weighting $\bar{S}_{l,i}$ for the regularization function \eqref{eq:insensitivity}. Here, we report the test set losses obtained by the continuous models (continuous lines) and the value for the overall $R$ function (dashed lines). We observe a very unstable test loss without insensitivity re-scaling for $R$ (magenta line). Hence, minimization with an overall $0.1$ re-scaling for $R$ is also shown (in cyan): in such case, the test loss on the continuous model remains low, but $R$ is extremely slowly minimized. Using the insensitivity re-weighting (in blue) proves to be a good trade-off between keeping the test set loss low and both minimizing $R$. This behavior is what we expected: 
the insensitivity re-weighting, acting parameter-wise (ie. there is a different value per each parameter), dynamically tunes the re-weighting of the overall regularization function $R$, allowing faster minimization with minimal or no performance loss. This is why we could use the same $\lambda_H$ and $\lambda_R$ values for all the simulations, despite optimizing different architectures on different datasets. Such robustness of the hyper-parameters over different dataset is a major practical strength of our approach. 
\section{Conclusion}
\label{sec:conclusion}

We presented HEMP, an entropy coding-based framework for compressing neural networks parameters. 
Our formulation efficiently estimates entropy beyond the first order 
and can be employed as regularizer to minimize the quantized parameters' entropy in gradient based learning, directly on the continuous parameters.
The experiments show that HEMP 
is not only an accurate proxy towards minimizing the entropy of the quantized parameters, but are also
pivotal to model the quantized parameters statistics and improve the efficiency of entropy coding schemes.
We also sided HEMP to LOBSTER, a state-of-the-art pruning strategy which introduces a prior on the weight's distribution which gives a further boost to the final model's compression.
Future works include the integration of a quantization technique designed specifically for deep models to HEMP.

\bibliography{main}

\begin{thebibliography}{10}
\expandafter\ifx\csname url\endcsname\relax
  \def\url#1{\texttt{#1}}\fi
\expandafter\ifx\csname urlprefix\endcsname\relax\def\urlprefix{URL }\fi
\expandafter\ifx\csname href\endcsname\relax
  \def\href#1#2{#2} \def\path#1{#1}\fi

\bibitem{samarakoon2019distributed}
S.~Samarakoon, M.~Bennis, W.~Saad, M.~Debbah, Distributed federated learning
  for ultra-reliable low-latency vehicular communications, IEEE Transactions on
  Communications.

\bibitem{iandola2016squeezenet}
F.~N. Iandola, S.~Han, M.~W. Moskewicz, K.~Ashraf, W.~J. Dally, K.~Keutzer,
  Squeezenet: Alexnet-level accuracy with 50x fewer parameters and< 0.5 mb
  model size, arXiv preprint arXiv:1602.07360.

\bibitem{sandler2018mobilenetv2}
M.~Sandler, A.~Howard, M.~Zhu, A.~Zhmoginov, L.-C. Chen, Mobilenetv2: Inverted
  residuals and linear bottlenecks, in: Proceedings of the IEEE conference on
  computer vision and pattern recognition, 2018, pp. 4510--4520.

\bibitem{molchanov2017variational}
D.~Molchanov, A.~Ashukha, D.~Vetrov, Variational dropout sparsifies deep neural
  networks, in: Proceedings of the 34th International Conference on Machine
  Learning-Volume 70, JMLR. org, 2017, pp. 2498--2507.

\bibitem{tartaglione2018learning}
E.~Tartaglione, S.~Leps{\o}y, A.~Fiandrotti, G.~Francini, Learning sparse
  neural networks via sensitivity-driven regularization, in: Advances in neural
  information processing systems, 2018, pp. 3878--3888.

\bibitem{louizos2017learning}
C.~Louizos, M.~Welling, D.~P. Kingma, Learning sparse neural networks through $
  l\_0 $ regularization, International Conference on Learning Representation
  (ICLR).

\bibitem{courbariaux2015binaryconnect}
M.~Courbariaux, Y.~Bengio, J.-P. David, Binaryconnect: Training deep neural
  networks with binary weights during propagations, in: Advances in neural
  information processing systems, 2015, pp. 3123--3131.

\bibitem{zhou2016dorefa}
S.~Zhou, Y.~Wu, Z.~Ni, X.~Zhou, H.~Wen, Y.~Zou, Dorefa-net: Training low
  bitwidth convolutional neural networks with low bitwidth gradients, arXiv
  preprint arXiv:1606.06160.

\bibitem{mishra2017wrpn}
A.~Mishra, E.~Nurvitadhi, J.~J. Cook, D.~Marr, Wrpn: wide reduced-precision
  networks, International Conference on Learning Representation (ICLR).

\bibitem{kim2015compression}
Y.-D. Kim, E.~Park, S.~Yoo, T.~Choi, L.~Yang, D.~Shin, Compression of deep
  convolutional neural networks for fast and low power mobile applications,
  International Conference on Learning Representation (ICLR).

\bibitem{xu2018deep}
Y.~Xu, Y.~Wang, A.~Zhou, W.~Lin, H.~Xiong, Deep neural network compression with
  single and multiple level quantization, in: Thirty-Second AAAI Conference on
  Artificial Intelligence, 2018.

\bibitem{wiedemann2019deepcabac}
S.~{Wiedemann}, H.~{Kirchhoffer}, S.~{Matlage}, P.~{Haase}, A.~{Marban},
  T.~{Marinc}, D.~{Neumann}, T.~{Nguyen}, H.~{Schwarz}, T.~{Wiegand},
  D.~{Marpe}, W.~{Samek}, Deepcabac: A universal compression algorithm for deep
  neural networks, IEEE Journal of Selected Topics in Signal Processing.

\bibitem{oktay2019scalable}
D.~Oktay, J.~Ball{\'e}, S.~Singh, A.~Shrivastava, Scalable model compression by
  entropy penalized reparameterization, arXiv preprint arXiv:1906.06624.

\bibitem{cheng2015exploration}
Y.~Cheng, F.~X. Yu, R.~S. Feris, S.~Kumar, A.~Choudhary, S.-F. Chang, An
  exploration of parameter redundancy in deep networks with circulant
  projections, in: Proceedings of the IEEE international conference on computer
  vision, 2015, pp. 2857--2865.

\bibitem{lee2021redundancy}
C.~Lee, Y.-B. Kim, H.~Ji, Y.~Lee, Y.~Hur, H.~Lim, On the redundancy in the rank
  of neural network parameters and its controllability, Applied Sciences 11~(2)
  (2021) 725.

\bibitem{zhang2018shufflenet}
X.~Zhang, X.~Zhou, M.~Lin, J.~Sun, Shufflenet: An extremely efficient
  convolutional neural network for mobile devices, in: Proceedings of the IEEE
  conference on computer vision and pattern recognition, 2018, pp. 6848--6856.

\bibitem{ullrich2017soft}
K.~Ullrich, E.~Meeds, M.~Welling, Soft weight-sharing for neural network
  compression, International Conference on Learning Representation (ICLR).

\bibitem{frankle2018lottery}
J.~Frankle, M.~Carbin, The lottery ticket hypothesis: Finding sparse, trainable
  neural networks, International Conference on Learning Representation (ICLR).

\bibitem{rastegari2016xnor}
M.~Rastegari, V.~Ordonez, J.~Redmon, A.~Farhadi, Xnor-net: Imagenet
  classification using binary convolutional neural networks, in: European
  conference on computer vision, Springer, 2016, pp. 525--542.

\bibitem{baldassi2018role}
C.~Baldassi, F.~Gerace, H.~J. Kappen, C.~Lucibello, L.~Saglietti,
  E.~Tartaglione, R.~Zecchina, Role of synaptic stochasticity in training
  low-precision neural networks, Physical review letters 120~(26) (2018)
  268103.

\bibitem{lin2017towards}
X.~Lin, C.~Zhao, W.~Pan, Towards accurate binary convolutional neural network,
  in: Advances in Neural Information Processing Systems, 2017, pp. 345--353.

\bibitem{shayer2017learning}
O.~Shayer, D.~Levi, E.~Fetaya, Learning discrete weights using the local
  reparameterization trick, International Conference on Learning Representation
  (ICLR).

\bibitem{belagiannis2018adversarial}
V.~Belagiannis, A.~Farshad, F.~Galasso, Adversarial network compression, in:
  Proceedings of the European Conference on Computer Vision (ECCV), 2018, pp.
  0--0.

\bibitem{han2015deep}
S.~Han, H.~Mao, W.~J. Dally, Deep compression: Compressing deep neural networks
  with pruning, trained quantization and huffman coding, International
  Conference on Learning Representation (ICLR).

\bibitem{witten1987arithmetic}
I.~H. Witten, R.~M. Neal, J.~G. Cleary, Arithmetic coding for data compression,
  Communications of the ACM 30~(6) (1987) 520--540.

\bibitem{seroussi1993lempel}
G.~Seroussi, A.~Lempel, Lempel-ziv compression scheme with enhanced adapation,
  uS Patent 5,243,341 (Sep.~7 1993).

\bibitem{pavlov2007lzma}
I.~Pavlov, Lzma sdk (software development kit) (2007).

\bibitem{preprint:lobster}
E.~Tartaglione, A.~Bragagnolo, A.~Fiandrotti, M.~Grangetto, Loss-based
  sensitivity regularization: towards deep sparse neural networks, arXiv
  preprint arXiv:2011.09905.

\bibitem{wang2019haq}
K.~Wang, Z.~Liu, Y.~Lin, J.~Lin, S.~Han, Haq: Hardware-aware automated
  quantization with mixed precision, in: Proceedings of the IEEE conference on
  computer vision and pattern recognition, 2019, pp. 8612--8620.

\bibitem{tung2018deep}
F.~Tung, G.~Mori, Deep neural network compression by in-parallel
  pruning-quantization, IEEE transactions on pattern analysis and machine
  intelligence 42~(3) (2020) 568--579.

\bibitem{tu2020pruning}
C.-H. Tu, J.-H. Lee, Y.-M. Chan, C.-S. Chen, Pruning depthwise separable
  convolutions for mobilenet compression, in: 2020 International Joint
  Conference on Neural Networks (IJCNN), IEEE, 2020, pp. 1--8.

\bibitem{he2019real}
Y.~He, Z.~Pan, L.~Li, Y.~Shan, D.~Cao, L.~Chen, Real-time vehicle detection
  from short-range aerial image with compressed mobilenet, in: 2019
  International Conference on Robotics and Automation (ICRA), IEEE, 2019, pp.
  8339--8345.

\bibitem{Wang2020APQ}
T.~Wang, K.~Wang, H.~Cai, J.~Lin, Z.~Liu, S.~Han, Apq: Joint search for nerwork
  architecture, pruning and quantization policy, in: Proceedings of the
  IEEE/CVF Conference on Computer Vision and Pattern Recognition, 2020.

\end{thebibliography}

\end{document}